\documentclass{article}

    \PassOptionsToPackage{numbers, compress}{natbib}

\usepackage[dblblindworkshop, final]{neurips_2025}

\workshoptitle{MATH-AI}

\usepackage[utf8]{inputenc} 
\usepackage[T1]{fontenc}    
\usepackage[hidelinks]{hyperref}       
\usepackage{url}            
\usepackage{booktabs}       
\usepackage{amsfonts}       
\usepackage{nicefrac}       
\usepackage{microtype}      
\usepackage{xcolor}         
\usepackage{multirow}
\usepackage{graphicx}
\usepackage{pifont}
\usepackage{subcaption}
\usepackage{nicefrac, xfrac}
\usepackage{svg}

\newcommand{\cmark}{\ding{51}}%
\newcommand{\xmark}{\ding{55}}%
\usepackage[acronym, style=super]{glossaries}
\newacronym{llm}{LLM}{Large Language Model}
\newacronym{lrm}{LRM}{Large Reasoning Model}
\newacronym{ai}{AI}{Artificial Intelligence}
\newacronym{cot}{CoT}{Chain-of-Thought}
\newacronym{rpm}{RPM}{Raven's Progressive Matrices}
\newacronym{sota}{SOTA}{state-of-the-art}
\newacronym{snr}{SNR}{signal-to-noise ratio}
\newacronym{vllm}{VLLM}{Visual Large Language Model}
\newacronym{pmf}{PMF}{probability mass function}

\definecolor{C1}{RGB}{255,147,0}
\definecolor{C2}{RGB}{0, 176, 80}
\definecolor{C3}{RGB}{112, 48, 160}

\title{I-RAVEN-X: Benchmarking Generalization and Robustness of Analogical and Mathematical Reasoning in Large Language and Reasoning Models}

\author{
Giacomo Camposampiero\\
IBM Research -- Zurich, ETH Zurich\\
{\tt\small giacomo.camposampiero1@ibm.com}
\And
Michael Hersche\\
IBM Research -- Zurich\\
{\tt\small michael.hersche@ibm.com}
\And
Roger Wattenhofer\\
ETH Zurich\\
{\tt\small wattenhofer@ethz.ch}
\And 
Abu Sebastian\\
IBM Research - Zurich\\
{\tt\small ase@zurich.ibm.com}
\And 
Abbas Rahimi\\
IBM Research -- Zurich\\
{\tt\small abr@zurich.ibm.com}
}

\begin{document}

\maketitle

\begin{abstract}
We introduce I-RAVEN-X, a symbolic benchmark designed to evaluate generalization and robustness in analogical and mathematical reasoning for \glspl*{llm} and \glspl*{lrm}. I-RAVEN-X extends I-RAVEN by increasing operand complexity, attribute range, and introducing perceptual uncertainty. 
Compared to \glspl*{llm}, empirical results on I-RAVEN-X show that \glspl*{lrm} achieve improved productivity and systematicity on longer reasoning relations and wider attribute ranges, respectively.
For instance, \glspl*{lrm} experience a significantly smaller degradation on arithmetic accuracy ($80.5\%\rightarrow 63.0\%$) compared to \glspl*{llm} ($59.3\%\rightarrow 4.4\%$).
However, \glspl*{lrm} are still significantly challenged by reasoning under uncertainty ($-61.8\%$ in task accuracy) and cannot effectively explore multiple probabilistic outcomes in superposition.
\end{abstract}

\section{Introduction}
Abstract reasoning is often regarded as a core feature of human intelligence. 
%
%
A wide range of benchmarks to assess abstract reasoning has been proposed in the past decade~\citep{bilker_development_2012,cherian_are_2023,chollet_measure_2019,niedermayr_rlp_2024}.
\gls*{rpm}~\citep{raven_ravens_1938,carpenter_what_1990}, a task associating vision with relational and analogical reasoning in a hierarchical representation, is one of the most prominent of them thanks to its extensive use to benchmark for abstract reasoning, analogy-making, and out-of-distribution (OOD) testing~\citep{benny_scale-localized_2021,hu_stratified_2021,malkinski_deep_2025,mitchell_abstraction_2021,zhang_raven_2019}.
RAVEN~\citep{zhang_raven_2019} represented the first attempts to build an automatically-generated dataset of \gls*{rpm} samples, enabling large-scale training of ML methods. 
I-RAVEN~\citep{hu_stratified_2021} improved RAVEN, proposing a new generation algorithm to avoid shortcut solutions that were possible in the original dataset.
However, RAVEN and I-RAVEN exhibit some limitations that hinder their reliability as benchmarks for reasoning proficiency in LLMs and LRMs. 
Firstly, most of the problems involve only a few operands, representing an overly simplistic subset of analogical and mathematical relations.
Most importantly, the test problems and their corresponding solutions are openly available online, increasing the risk of potential data leakage from the model's pre- and post-training stages as previously observed in other settings~\citep{mirzadeh_gsm-symbolic_2025}.
Furthermore, assuming the availability of an \emph{oracle perception} has become a standard practice in their translation from visual to textual (symbolic) tasks (necessary to test language-only models)~\citep{webb_emergent_2023, hu_-context_2023}.
This assumption is reasonable when the scope of the investigation is limited to the reasoning component; however, it falls short when we zoom out to more complex, end-to-end systems, as it bypasses crucial steps of the original visual analogical reasoning, such as filtering irrelevant attributes and accounting for the uncertainty of the perception module.

To tackle these problems, this paper makes the following contributions:
\begin{enumerate}
    \item introduces I-RAVEN-X, an enhanced, symbolic version of the standard I-RAVEN benchmark that enables testing the generalization and robustness to simulated perceptual uncertainty in text-based language and reasoning models (see Figure \ref{fig:iravenx}),
    \item highlights that \glspl*{lrm} consistently generalize better than \glspl*{llm} in terms of productivity and systematicity, but significantly fail to reason under uncertainty.
\end{enumerate}

\section{Methods}

\subsection{I-RAVEN-X: testing generalization and robustness of reasoning in LLMs and LRMs}
\label{sec:dataset}

\begin{figure}[b!]
    \centering
    \includegraphics[width=\linewidth]{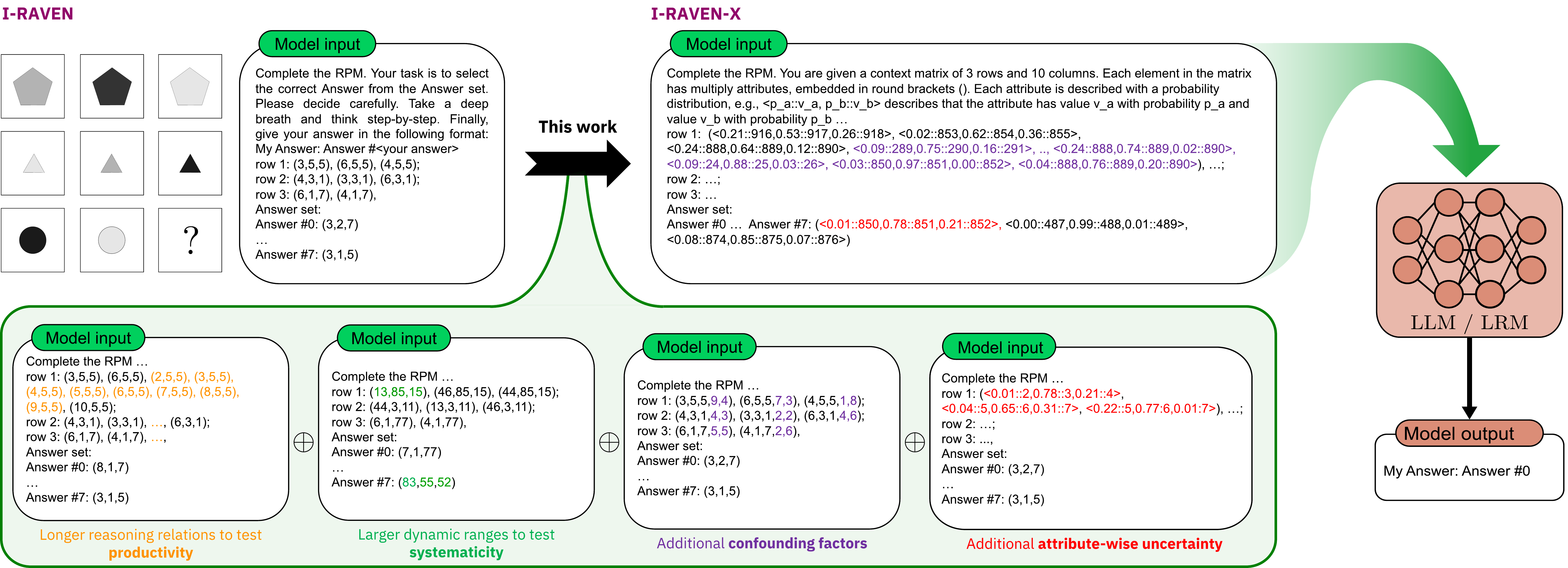}
    \caption{This figure highlights all the different axes of generalization and robustness to uncertainty, which I-RAVEN-X stresses. Compared to standard I-RAVEN \textbf{(a)}, I-RAVEN-X \textbf{(b)} involves more panels per row (10 vs. 3) and larger attribute dynamic ranges (up to $100\times$ more values per attribute). In addition, it is possible to introduce uncertainty in the reasoning process through confounders (such as panels' background and color patterns within objects) and smoothening the attribute values' distributions (displayed on the right for the panel in position $(1,10)$) \textbf{(c)}.
    We adopt a visual representation of the panels and their attributes for clarity of explanation; in practice, however, our dataset is purely symbolic and has not been extended yet to the visual domain.}
    \label{fig:iravenx}
\end{figure}

We propose a fully-symbolic, parametrizable dataset to evaluate LLMs and LRMs, dubbed I-RAVEN-X.
Some examples from the dataset are included in Figure \ref{fig:iravenx}.
I-RAVEN-X enhances the original I-RAVEN (more extensively described in Appendix \ref{app:iraven}) over different dimensions:
\begin{enumerate}
    \item \textbf{Productivity}: we parametrize the number of operands in the reasoning relations (e.g., using 3$\times$10 matrices instead of 3$\times$3, \fcolorbox{black}{C1}{\rule{0pt}{4pt}\rule{4pt}{0pt}} in Figure \ref{fig:iravenx});
    \item \textbf{Systematicity}: we introduce larger dynamic ranges for the operand values (e.g., 1000 attribute values instead of 10, \fcolorbox{black}{C2}{\rule{0pt}{4pt}\rule{4pt}{0pt}} in Figure \ref{fig:iravenx});
    \item \textbf{Robustness to confounding factors}: we augment the set of original attributes in RPM with randomly sampled values, which do not contribute to the reasoning (\fcolorbox{black}{C3}{\rule{0pt}{4pt}\rule{4pt}{0pt}} in Figure \ref{fig:iravenx});
    \item \textbf{Robustness to non-degenerate value distributions}: we smoothen the distributions of the input values corresponding to the generative factors (\fcolorbox{black}{red}{\rule{0pt}{4pt}\rule{4pt}{0pt}} in Figure \ref{fig:iravenx}).
\end{enumerate}
Practically speaking, 1. and 2. enable testing the generalization of LLMs and LRMs to longer reasoning chains and an increased number of concepts.
On the other hand, 3. and 4. allow to loosen the strong assumption of an oracle perception, simulating an imperfect sensory front-end while operating with text-only language models, hence providing an additional focus on the robustness of reasoning under uncertainty.
More details on the design of $1-4$ are included in Appendix \ref{app:details}.
In addition, the original I-RAVEN was narrowed down to a single constellation (\texttt{center}, containing a single object per panel), which was observed to be simultaneously a strong test for a wide range of logical and arithmetic skills and unexpectedly challenging.

\subsection{Models and prompting techniques}
\label{sec:models}
We focus our study on two \gls*{sota} \glspl*{lrm} (the closed-source OpenAI o3-mini model and the open-source DeepSeek R1 model~\citep{deepseek-r1-short}, together with its distilled version based on Llama 70B) and \glspl*{llm} (the proprietary GPT-4~\citep{openai_gpt-4_short} and the open-source Llama-3 70B~\citep{llama_short}).
A more precise accounting of the version of the model used, along with additional details on the prompting engineering techniques adopted in our experiments, is presented in Appendix~\ref{app:models_prompts}.
In Appendix~\ref{appx:o1}, we include an additional comparison between o3-mini and its predecessor, o1.

\section{Results}
In this section, we evaluate the generalization and robustness of the analogical and arithmetic reasoning capabilities of \glspl*{lrm} and \glspl{llm} using I-RAVEN and I-RAVEN-X.
In particular, we aim to answer the following research questions: How well do the analogical and mathematical capabilities of LLMs and LRMs generalize in terms of productivity and systematicity \textbf{(Q1)}? How robust are LLMs and LRMs when confronted with reasoning under uncertainty \textbf{(Q2)}?

\subsection{LRMs are stronger analogical and mathematical reasoners than LLMs}
\label{sec:comparison}
To answer Q1, we benchmark the productivity and systematicity of the models introduced in Section \ref{sec:models} on I-RAVEN and I-RAVEN-X.
Table \ref{tab:lrmresults} reports the results of the evaluation.
We observe that \glspl*{lrm} are much stronger reasoners than \glspl*{llm} when challenged with the longer reasoning rules and attribute ranges in I-RAVEN-X, especially when we consider mathematical reasoning (additive relations).
While \glspl*{llm} show a massive drop in arithmetic accuracy on I-RAVEN-X, nearing $0\%$ for comparable prompt complexity, \glspl*{lrm} are affected by a much smaller arithmetic degradation on average.
These marked gains in arithmetic reasoning performance, with improvements reaching up to 65.4\% in certain settings, suggest that \glspl*{lrm} can more comfortably identify and generalize (in productivity and systematicity) mathematical rules compared to \glspl*{llm}.

Moreover, we can also see that \glspl*{lrm} achieve reasoning accuracy on par with \glspl*{llm} despite using significantly less engineered prompts; when the prompt complexity is comparable, on the other hand, \glspl*{lrm} consistently outperform \glspl*{llm} on the investigated benchmarks.
o3-mini, for instance, shows no drops in accuracy on I-RAVEN-X and a $6\%$ drop on I-RAVEN compared to GPT-4 while using only \sfrac{1}{21} of the prompts.
When we compare the same two models on similar prompt complexities (that is, using entangled prompting in both settings, but still retaining a 1:7 ratio between \glspl*{lrm} and \glspl*{llm} due to self-consistency), o3-mini emerges as a clear winner, showing an average $6.5\%$ increase in accuracy.
These results suggest that \glspl*{lrm} do not require incorporating as much inductive bias (through prompt engineering) as \glspl*{llm} do, and that the production of ``thinking'' tokens generally makes the reasoning process more robust.

\begin{table}[h!]
\resizebox{\textwidth}{!}{
\begin{tabular}{lccccccccccc}
\toprule
\multirow{4}{*}{\textbf{Model}} & \multirow{4}{*}{\textbf{ICL}}  & \multirow{4}{*}{\textbf{Prompts}} & \multicolumn{3}{c}{\textbf{I-RAVEN} (3$\times$3)} & \multicolumn{6}{c}{\textbf{I-RAVEN-X} (3$\times$10)} \\
\cmidrule(r){4-6}\cmidrule(r){7-12}
& & &\multicolumn{3}{c}{\textbf{Range 10}}  & \multicolumn{3}{c}{\textbf{Range 100}}      & \multicolumn{3}{c}{\textbf{Range 1000}}    \\
\cmidrule(r){4-6}\cmidrule(r){7-9}\cmidrule(r){10-12}
& & & Task & Arithm. & Tok. &  Task & Arithm.  & Tok. & Task & Arithm.  & Tok. \\
\cmidrule(r){1-3}\cmidrule(r){4-6}\cmidrule(r){7-9}\cmidrule(r){10-12}
Llama-3 70B      & \cmark & 21 &  85.0 & 45.0 & 21 & 73.0 &  2.6  & 21 & 74.2 & 0.4 & 21 \\
GPT-4             & \xmark & 21 &   93.2 & 73.6 & 21  & 79.6 & 25.1 & 21 &  76.6 & 8.4 & 21 \\
Llama-3 70B       & \cmark & 7  &   79.0 & 31.0  & 21 & 72.6 &  0.0 & 21 & 74.0 & 0.4 & 21 \\
GPT-4             & \xmark & 7  & 74.8 & 27.2  & 21 & 72.8 & 2.7  & 21 & 74.0 & 1.1 & 21 \\
\cmidrule(r){1-3}\cmidrule(r){4-6}\cmidrule(r){7-9}\cmidrule(r){10-12}
OpenAI o3-mini (med.)& \xmark &1  &  86.6 &  74.4 & 5445 &  77.6 & 53.2 & 7884 & 81.0 & 60.8 & 7209 \\
OpenAI o3-mini (high) & \xmark &1  & 92.6  &  86.1 & 9867 &  82.4 & 63.5 & 19041 & 80.6 & 60.1 & 19449 \\
DeepSeek R1    & \xmark & 1 &  80.6 & 74.8 & 4486 & 84.0 & 67.7 & 5550 & 82.8 & 65.8 &5505 \\
DeepSeek R1 dist. & \xmark & 1&  78.4 & 69.4 & 5192  & 67.0 & 52.9 & 6690 & 72.0 & 54.4 &  6324\\
\bottomrule
\end{tabular}
}
\caption{Full task accuracy (\% of test examples correctly predicted) and arithmetic accuracy (\% of the attributes in the test examples governed by an arithmetic relation correctly predicted) of \glspl*{llm} and \glspl*{lrm} on I-RAVEN and I-RAVEN-X. We report if In-Context Learning (ICL) examples of the task were added to the prompt, the number of total prompts fed into the model (some techniques, such as self-consistency and disentangled prompting require querying the model multiple times), and the number of tokens generated by the model. ``Range'' indicates the dynamic range of the attributes' values. ``Tok.'' indicates the average number of output tokens of the model.}
\label{tab:lrmresults}
\end{table}

\subsection{LRMs are significantly challenged by reasoning under uncertainty}

The results in Section~\ref{sec:comparison} show that \glspl*{lrm} can solve analogical and mathematical reasoning tasks more accurately than \glspl*{llm}.
However, would they be capable of retaining the same robustness in scenarios where uncertainty is introduced (Q2)?
To answer this question, we benchmark two LRMs with I-RAVEN-X with uncertainty as proposed in Section \ref{sec:dataset}.
Due to the failures shown in the previous section, we do not consider LLMs for these experiments.
The empirical results of this study are reported in Table \ref{tab:uncertainty}.

Firstly, we observe that \glspl*{lrm} perform significantly worse when noise factors that simulate perceptual uncertainty are integrated into the experiments.
For instance, o3-mini's accuracy dropped by $11.2\%$ and $15.2\%$ on task and arithmetic accuracy, respectively, when evaluated with 10 additional confounding attributes.
R1, on the other hand, is more robust to confounders ($5.8\%$ and $12.2\%$ drops on task and arithmetic accuracy), but it performs much worse when the attribute values' distributions are smoothened, losing up to $19.8\%$ of task accuracy in the harshest scenario.
o3-mini shows a much smaller degradation ($5.4\%$) in this setting.

When both the confounders and distribution smoothening are evaluated together at their maximum level, we observe a sharp drop in task accuracy for both o3-mini (to $17.0\%$)  and DeepSeek R1 (to $22.8\%$), bringing them close to random chance ($12.5\%$). 
Different causes could play a role in this significant degradation: an increasing complexity of the prompts, which might impair the model's ability to detect and mimic patterns in the input, or a more general limitation in modeling probabilistic reasoning that requires maintaining coherence across multiple uncertain variables.

\begin{table}[h!]
\centering
\resizebox{\linewidth}{!}{
     \begin{tabular}{llccccccc}
     \toprule
      &  & & \multicolumn{3}{c}{\textbf{OpenAI o3-mini}} & \multicolumn{3}{c}{\textbf{DeepSeek R1}}  \\
     \textbf{Experiment} &\textbf{Confounders\,(SNR)} &$\mathbf{p_L}$ &  {Task} &  {Arith.} & {Tokens} & {Task} &  {Arith.} & Tokens \\
     \cmidrule(r){1-3}\cmidrule(r){4-6}\cmidrule(r){7-9}
     & 0 ($\infty$) & 1.00 & 81.0 & 60.8 & 7209 & 82.8  & 65.8 & 6324  \\
     \cmidrule(r){1-3}\cmidrule(r){4-6}\cmidrule(r){7-9}
     \multirow{4}{*}{\textbf{(a)}} & 1 (4.77) & 1.00 & 76.0 & 53.2 & 11521 &  78.2 & 55.2 & 8919  \\
     &3 (0.00) & 1.00 & 75.6 & 51.7 & 11669 &  80.2 & 58.2 & 8429   \\
     &5 ($-2.22$) & 1.00  & 71.2 & 48.3 & 12640 &  78.6 & 55.9 & 8681  \\
     &10 ($-5.23$)& 1.00 & 69.8 & 45.6 & 13709 &  77.0 & 53.6 & 8912  \\
    \cmidrule(r){1-3}\cmidrule(r){4-6}\cmidrule(r){7-9}
     \multirow{2}{*}{\textbf{(b)}}&0 ($\infty$)& 0.70 & 75.0 & 51.7 & 13112 &  67.4 & 44.9 & 6995  \\
     &0 ($\infty$)& 0.51 & 75.6  & 53.2 & 13028 & 63.0  & 46.4 & 7518  \\
    \cmidrule(r){1-3}\cmidrule(r){4-6}\cmidrule(r){7-9}
     \textbf{(c)}&10 ($-5.23$)& 0.51 & 17.0 & 41.1  & 18482 & 23.2  & 45.3 &  7147\\
     \bottomrule
     \end{tabular}
}
\caption{Task and arithmetic accuracy (\%) on I-RAVEN-X (range [0,1000]) with different numbers of confounders, from 0 (no confounders, SNR=$\infty$) to 10 (SNR=$-5.23$ dB), and different attributes' distribution smoothening (bin-smoothening strategy, with different probabilities assigned to the correct value bin $p_L$). 
We show experiments with: a) only confounders; b) only the attributes' distribution smoothing; c) both confounders and distribution smoothing. 
We report the number of output tokens to quantify the {reasoning effort} adopted on average by the model to find a solution.}
\label{tab:uncertainty}
\end{table} 

\section{Conclusion}
This work presents I-RAVEN-X, a novel, symbolic benchmark for testing the generalization and robustness of analogical and mathematical reasoning.
I-RAVEN-X is then used to evaluate these capabilities in LLMs and LRMs.
Compared to \glspl*{llm}, \glspl*{lrm} achieve improved productivity and systematicity on longer reasoning relations and wider attribute ranges, respectively.
For instance, \glspl*{lrm} experience a significantly smaller degradation on arithmetic accuracy ($80.5\%\rightarrow 63.0\%$) compared to \glspl*{llm} ($59.3\%\rightarrow 4.4\%$).
However, LRMs are still significantly challenged by reasoning under uncertainty ($-61.8\%$ in task accuracy) and cannot explore multiple probabilistic outcomes at the same time.
One limitation of our work consists of exploring the causal relationship between reasoning under uncertainty, prompt efficiency, and reasoning accuracy; further investigating this relation is left for future work.
The dataset and the experiments' code are available at \url{https://github.com/IBM/raven-large-language-models}.
This paper summarizes the contributions presented in two recent publications, \citet{hersche_towards_2025} and \citet{camposampiero_2025_can}.

\newpage
\appendix

\section{I-RAVEN dataset}
\label{app:iraven}
\begin{figure}[h!]
    \centering
    \includegraphics[width=\linewidth]{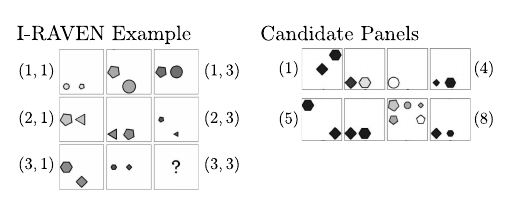}
    \caption{RPM example from I-RAVEN.}
    \label{fig:iraven}
\end{figure}

Raven’s progressive matrices (RPM) is a visual task that involves perceiving pattern continuation and elemental abstraction as well as deducing relations based on a restricted set of underlying rules in a process that mirrors the attributes of advanced human intelligence.
In this work, we focus on the I-RAVEN dataset. 
Each RPM test in I-RAVEN is an analogy problem presented as a $3\times 3$ pictorial matrix of context panels. 
Every panel in the matrix is filled with several geometric objects based on a certain rule, except the bottom-right panel, which is left blank.
Figure~\ref{fig:iraven} includes an I-RAVEN example test. 
The task is to complete the missing panel by picking the correct answer from a set of (eight) candidate answer panels that match the implicit generation rule on every attribute. 
The object's attributes (color, size, shape, number, position) are governed by individual underlying rules: 
\begin{itemize}
    \item \textit{constant}, the attribute value does not change per row;
    \item \textit{arithmetic}, the attribute value of the third panel corresponds to either the sum or the difference of the first two panels of the row;
    \item \textit{progression}, the attribute value monotonically increases or decreases in a row by 1 or 2;
    \item \textit{distribute three}, the set of the three different values remains constant across rows, but the individual attribute values get shifted to the left or to the right by one position at every row; it also holds column-wise.
\end{itemize}
Each panel contains a variable number of objects (minimum one, maximum nine) arranged according to one of seven different constellations (center, distribute-four, distribute-nine, 
left-right, up-down, in-out-center, and in-out-four). 

\section{I-RAVEN-X implementation details}
\label{app:details}

\subsection{Productivity and Systematicity}
Our new benchmark maintains I-RAVEN's four rules and three attributes but allows for a parameterizable number of columns ($g$) and a dynamic range of attribute values ($m$). 
When generating a new \gls*{rpm} example, we uniformly sample from one of the available rules (\texttt{constant}, \texttt{progression}, \texttt{arithmetic}, and \texttt{distribute three}). 
Note that the attribute \texttt{shape} does not incur the \texttt{arithmetic} rule. 

In the following, we describe the generation process of the RPM context matrix of size $3\times g$ for the individual rules. 
The overall goal is that the values stay in the range $[ 0, m-1]$. 
\begin{itemize}
\item \texttt{constant}: For each row, we uniformly sample an integer from the set $\lbrace 0, 1, ..., m-1 \rbrace$, and duplicate along the row. 
\sloppy
\item \texttt{progression}: First, we uniformly sample the progressive increment/decrement ($\delta$) from the set $\lbrace -2, -1, +1, +2 \rbrace$. In case of a positive increment, we first define the values of the right-most columns, by uniformly sampling from the set $\lbrace (g-1)\cdot \delta, ..., m-1  \rbrace$ for each row.
Then, the rest of the matrix is completed by applying the progression rule. 
The sampling for a negative $\delta$ is done specularly from the first column. 
\item \texttt{arithmetic}: The attribute values of the first $g-1$ panels are either added (\texttt{arithmetic plus}) or subtracted (\texttt{arithmetic minus}), yielding the attribute value of the last panel in the row. 
In \texttt{arithmetic plus}, we sequentially sample the values from the first $g-1$ panels in the row. For each panel, we set the sampling range to $\lbrace 0, ..., m - s \rbrace$, where $s$ is the sum of the already sampled panels in the row. 
Afterward, the first $g-1$ panels are shuffled. 
Finally, the values of the last panels are the sum of the first $g-1$ ones, applied row-wise. 
For \texttt{arithmetic minus}, we apply the same sampling strategy but leave the first column empty. 
The value of the first column is then defined as the sum of the other columns.
\item \texttt{distribute-n}: We uniformly sample distinct values for the first row from $\lbrace 0, ..., m-1  \rbrace$. 
The content of the remaining rows is defined by applying a circular shift per row (either right or left). 
\end{itemize}

Finally, we generate the candidate answers using I-RAVEN's attribute bisection tree~\cite{hu_stratified_2021}. 
The original RAVEN dataset had a flaw in the generation of the answer set. 
Each distractor in the answer set (i.e., a wrong answer candidate) was generated by randomly altering one attribute of the correct answer.
As a result, one could predict the correct answer by taking the mode of the answer candidates without looking at the context matrix, therefore bypassing the actual reasoning task. 
As a remedy, the attribute bisection tree generates unbiased answers that are well-balanced.

\subsection{Confounding attributes}
Confounding attributes represent properties and patterns that can be extracted from the visual inputs by a front-end perception module but are not relevant to the reasoning process.
This could be the case, for instance, when the attributes are extracted by unsupervised vision models such as Variational Autoencoders~\citep{kingma_auto-encoding_2013} or even a multi-modal \gls*{llm} that is prompted to extract the attributes.
In Figure~\ref{fig:iravenx}, confounding attributes are represented by the background of the input panels and the color patterns, which sometimes appear inside the objects.
In I-RAVEN-X, we integrate confounders by extending the set of original attributes of each panel with an arbitrary number of additional attributes uniformly sampled in the interval $[0, m-1]$, where $m$ is the range of the attributes' values.
For large enough $m$, the probability of sampling values that fit a valid rule is negligible, and hence, confounders do not introduce ambiguities in the choice of the answer panel.
However, they linearly reduce the \gls*{snr} in the reasoning process, requiring models to implement strategies to filter out noisy input components.

\subsection{Smooth attribute values' distributions}
We deviate from the original I-RAVEN degenerate attributes' distributions and introduce variance, which allows us to test the robustness of the models when reasoning with uncertain attribute values.
In practice, we smoothen the original attributes' distributions using a three-bins strategy, where the probability of the true value $T$ is $p(T)\sim \mathcal{U} (p_L, 1), p_L>0.5$ and the probabilities of its two neighboring values are $p(N_1)\sim \mathcal{U} (0, 1-p(T))$ and $p(N_2)=1-p(T)-p(N_1)$.
Note that the motivation behind the three-bins strategy is to introduce variance with minimal additional cost for \glspl*{lrm}' prompt complexity.

\newpage
\section{Models and prompting details}
\label{app:models_prompts}
\paragraph{LLMs}
We focused our evaluations on text-only \glspl*{llm}. 
There exist attempts~\cite{mitchell_comparing_2024,jiang_marvel_2024, cao_what_2024, ahrabian_curious_2024, zhang_how_2024} that leverage vision support of some multi-modal \glspl*{llm} (e.g., GPT-4V) directly feeding the models with visual \gls*{rpm} data; however, they achieve consistently lower reasoning performance than with text-only prompting. 
The \gls*{sota} \gls*{llm}-based abstract reasoning approach~\cite{hu_-context_2023} relied on reading out GPT-3's (\texttt{text-davinci-002}) token probabilities.
However, this model is no longer accessible to users and its successive iterations do not allow the retrieval of prediction logits.
Hence, we considered discrete classification approaches that are based on output strings rather than distribution over tokens.
In particular, we investigated two \gls*{sota} \glspl*{llm}: the proprietary GPT-4~\cite{openai_gpt-4_short}\footnote{GPT-4 was accessed between 07/03/2024--10/30/2024.} (\texttt{gpt-4-0613}) and the open-source Llama-3 70B~\cite{llama_short}\footnote{The model weights were downloaded and evaluated locally. Unless stated otherwise, we use the base model without instruction tuning.}.
During initial tests, GPT-4o yielded worse results than GPT-4, hence we focused on GPT-4. 
Different prompting engineering techniques were integrated to improve the overall accuracy of these models:
\begin{itemize}
    \item \textbf{Disentangled prompting}, a compositionally structured approach that queries the \gls*{llm} for individual attribute prediction. Disentangled prompting simplifies the task, but increases the number of queries by 3$\times$ (where 3 is, in this case, the number of attributes). In our experiments, disentangled prompts were only used for some experiments in LLMs, increasing the number of prompts from 7 to 21. 
    \item \textbf{Self-consistency}~\citep{wang_self-consistency_2023, lewkowycz_solving_2022}, which consists in querying the model multiple times ($n=7$ times), sampling the next token from the distribution with a non-zero soft-max temperature. We find the optimal soft-max temperature for GPT-4 ($T=0.5$) and Llama-3 70 B ($T=0.4$) via a grid search on a subset of 50 I-RAVEN problems. 
    We did not explore the effect of other parameters, such as top-k or top-p, and set them to the default values. 
    The final prediction is determined by a majority vote over the sampled outputs.  
    The selection of an odd number of samples (i.e., $n=7$) helps to prevent potential ties. 
    \item \textbf{In-context learning}: for a better understanding of the RPM task, we optionally prefix 16 in-context examples to the prompt~\cite{brown_language_2020}. In the predictive classification approach (where no answer candidates are provided), we simply provide complete example RPM matrices. The in-context samples are randomly selected from I-RAVEN's training set. Examples that had the same context matrix as the actual task are discarded and re-sampled to prevent shortcut solutions. 
\end{itemize}

\paragraph{LRMs}
For the OpenAI o3-mini model, we use the \texttt{o3-mini-2025-01-31} via the OpenAI API. By default, reasoning efforts were set to \texttt{medium} and the number of reasoning tokens to 25,000.
For DeepSeek R1 model, the full model with 671B parameters was serviced by \url{www.together.ai}, whereas the distilled version was run locally on 8 NVIDIA A100 GPUs. The maximum number of reasoning tokens was set to 25,000, the temperature to $0.6$, and top-p to $0.7$.
Self-consistency~\citep{wang_self-consistency_2023, lewkowycz_solving_2022} and attributes' scaling~\citep{hu_-context_2023} were not used in experiments with LRMs. 
Moreover, no in-context examples of the tasks~\citep{brown_language_2020} were provided since they were previously observed to be hurtful for \glspl*{lrm}~\citep{deepseek-r1-short}.
We also restrict the investigation to a subset of 500 randomly sampled \gls*{rpm} tests in both I-RAVEN and I-RAVEN-X (due to budget constraints), which we observed to be representative enough of the entire test set~\citep{hu_-context_2023}. 
We report some examples of the prompts used in our experiments in Tables \ref{fig:iraven_prompt}, \ref{fig:iravenx_prompt}, \ref{fig:iravenx_prompt_conf}, and \ref{fig:iraven_prompt_unc}.
The prompting style for embracing CoT was inspired by~\cite{wust_bongard_2024}. 
For automatic retrieval of the model's answer, we prompt it to provide its answer in the format ``My Answer: Answer \#<your answer>''. 
By default, answer panel \#0 is predicted if no answer can be retrieved. 
Contrary to LLMs, all the empirical results reported for LRMs are obtained using entangled prompts.

\newpage
Some examples of I-RAVEN and I-RAVEN-X prompts used for LRMs are reported in the following Tables.

\begin{table}[h!]
    \centering
    \begin{tabular}{p{\linewidth}}
        \toprule
        \vspace{-2mm}
        Complete the Raven's progressive matrix. Your task is to select the correct Answer from the Answer set. Please decide carefully. Take a deep breath and think step-by-step. Finally, give your answer in the following format: My Answer: Answer \#<your answer>\\
        \vspace{-1mm}
        row 1: (3,5,5), (6,5,5), (4,5,5); \\
        row 2: (4,3,1), (3,3,1), (6,3,1); \\
        row 3: (6,1,7), (4,1,7), \\
        \vspace{-1mm}
        Answer set: \\
        \qquad Answer \#0: (3,2,7)\\
        \qquad Answer \#1: (7,1,5)\\
        \qquad Answer \#2: (7,2,5)\\
        \qquad Answer \#3: (7,2,7)\\
        \qquad Answer \#4: (7,1,7)\\
        \qquad Answer \#5: (3,1,7)\\
        \qquad Answer \#6: (3,2,5)\\
        \qquad Answer \#7: (3,1,5) \\
    \bottomrule
    \end{tabular}
    \caption{Example prompt for an I-RAVEN task.}
    \label{fig:iraven_prompt}
\end{table}

\begin{table}[h!]
    \centering
    \begin{tabular}{p{\linewidth}}
        \toprule
        \vspace{-2mm}
    Complete the Raven's progressive matrix. Your task is to select the correct Answer from the Answer set. Please decide carefully. Take a deep breath and think step-by-step. Finally, give your answer in the following format: My Answer: Answer \#<your answer> \\
    \vspace{-1mm}
    \footnotesize
    row 1: (6,16,9), (7,15,9), (70,14,9), (93,13,9), (88,12,9), (77,11,9), (83,10,9), (22,9,9), (39,8,9), (27,7,9); \\
    \footnotesize
    row 2: (7,12,24), (70,11,24), (93,10,24), (88,9,24), (77,8,24), (83,7,24), (22,6,24), (39,5,24), (27,4,24), (6,3,24); \\
    \footnotesize
    row 3: (70,35,52), (93,34,52), (88,33,52), (77,32,52), (83,31,52), (22,30,52), (39,29,52), (27,28,52), (6,27,52), \\
    \vspace{-1mm}
    Answer set:\\
    \qquad Answer \#0: (7,26,52) \\
    \qquad Answer \#1: (83,55,52)\\
    \qquad Answer \#2: (7,26,37)\\
    \qquad Answer \#3: (83,55,37)\\
    \qquad Answer \#4: (7,55,52)\\
    \qquad Answer \#5: (83,26,37)\\
    \qquad Answer \#6: (7,55,37)\\
    \qquad Answer \#7: (83,26,52)\\
    \bottomrule
    \end{tabular}
    \caption{Example prompt for an I-RAVEN-X task.}
    \label{fig:iravenx_prompt}
\end{table}

\begin{table}[h!]
    \centering
    \tiny
    \begin{tabular}{p{\linewidth}}
        \toprule
        \vspace{-2mm}
    Complete the Raven's progressive matrix. Your task is to select the best matching Answer from the Answer set. Please decide carefully. Take a deep breath and think step-by-step. Finally, give your answer in the following format: My Answer: Answer \#<your answer> \\
    \vspace{-1mm}
    row 1: (917,854,889,837,449,40,616,988,225,603,813,154,860), (290,853,889,310,920,885,291,416,926,503,379,786,859), (532,852,889,336,540,95,33,182,41,215,990,859,625), (25,851,889,948,465,970,253,795,956,622,323,735,535), (31,850,889,846,149,643,802,187,413,101,300,378,181), (43,849,889,700,975,580,488,662,820,977,189,160,955), (574,848,889,484,18,951,173,279,247,567,639,939,730), (761,847,889,971,245,547,175,991,94,306,976,778,188), (576,846,889,547,182,955,995,410,545,537,859,368,146), (291,845,889,544,515,965,647,155,660,835,167,363,578); \\
    row 2: (290,898,875,416,729,621,255,121,775,992,332,824,69), (532,897,875,617,602,91,626,959,328,566,572,496,129), (25,896,875,507,14,482,3,638,723,822,326,152,311), (31,895,875,551,141,165,894,867,142,856,245,396,325), (43,894,875,645,712,987,788,382,795,149,295,457,63), (574,893,875,269,762,290,698,804,252,56,328,850,702), (761,892,875,621,590,319,785,4,122,627,517,924,88), (576,891,875,268,299,764,678,718,860,626,845,523,1), (291,890,875,860,69,712,754,590,214,674,171,773,227), (917,889,875,802,908,433,515,585,256,102,529,939,585); \\
    row 3: (532,497,831,73,406,82,149,646,932,466,196,966,172), (25,496,831,76,880,109,467,76,845,392,673,736,51), (31,495,831,79,825,847,494,174,270,472,649,164,234), (43,494,831,39,960,182,917,180,643,977,698,321,467), (574,493,831,553,583,258,422,840,680,109,870,539,289), (761,492,831,481,548,81,43,180,359,410,733,702,708), (576,491,831,882,329,883,287,624,816,453,120,316,349), (291,490,831,398,434,521,426,600,224,181,827,281,512), (917,489,831,611,791,841,260,28,125,408,122,577,903), \\
    \vspace{-1mm}
    Answer set:\\
    Answer \#0: (290,488,875,657,175,669,825,660,980,305,71,297,764)\\
    Answer \#1: (851,488,875,785,95,663,714,937,607,543,958,80,215)\\
    Answer \#2: (290,451,831,808,72,151,7,665,312,920,665,806,177)\\
    Answer \#3: (290,488,831,340,114,819,129,10,922,744,948,540,925)\\
    Answer \#4: (851,451,875,714,337,713,987,115,520,218,644,222,463)\\
    Answer \#5: (851,488,831,948,251,490,394,977,846,124,951,827,501)\\
    Answer \#6: (290,451,875,761,816,59,950,670,732,542,237,552,272)\\
    Answer \#7: (851,451,831,9,552,304,979,949,86,118,847,82,575) \\
    \bottomrule
    \end{tabular}
    \caption{Example prompt for the I-RAVEN-X task with confounders.}
    \label{fig:iravenx_prompt_conf}
\end{table}

\begin{table}[h!]
    \centering
    \tiny
    \begin{tabular}{p{\linewidth}}
        \toprule
        \vspace{-2mm}
        Complete the Raven's progressive matrix. You are given a context matrix of 3 rows and 10 colums. Each element in the matrix has multiply attributes, embedded in round brackets (). Each attribute is described with a probability distribution, e.g., <p\_a::v\_a, p\_b::v\_b> describes that the attribute has value v\_a with probability p\_a and value v\_b with probability p\_b. Your task is to select the best matching Answer from the Answer set. Please decide carefully. Take a deep breath and think step-by-step. Finally, give your answer in the following format: My Answer: Answer \#<your answer>\\
        \vspace{-1mm}
        row 1: (<0.21::916,0.53::917,0.26::918>, <0.02::853,0.62::854,0.36::855>, <0.24::888,0.64::889,0.12::890>), (<0.09::289,0.75::290,0.16::291>, <0.12::852,0.74::853,0.14::854>, <0.11::888,0.85::889,0.04::890>), (<0.44::531,0.55::532,0.01::533>, <0.36::851,0.63::852,0.01::853>, <0.24::888,0.74::889,0.02::890>), (<0.09::24,0.88::25,0.03::26>, <0.03::850,0.97::851,0.00::852>, <0.04::888,0.76::889,0.20::890>), (<0.08::30,0.58::31,0.34::32>, <0.02::849,0.97::850,0.01::851>, <-0.00::888,0.91::889,0.09::890>), (<0.20::42,0.51::43,0.29::44>, <0.01::848,0.97::849,0.02::850>, <0.25::888,0.70::889,0.05::890>), (<0.12::573,0.87::574,0.01::575>, <0.06::847,0.78::848,0.16::849>, <0.01::888,0.99::889,0.00::890>), (<0.04::760,0.82::761,0.14::762>, <0.08::846,0.70::847,0.22::848>, <0.04::888,0.77::889,0.19::890>), (<0.04::575,0.54::576,0.42::577>, <0.46::845,0.54::846,-0.00::847>, <0.01::888,0.91::889,0.08::890>), (<0.15::290,0.85::291,0.00::292>, <0.04::844,0.78::845,0.18::846>, <0.30::888,0.66::889,0.04::890>);\\
        row 2: (<0.01::289,0.81::290,0.18::291>, <0.19::897,0.59::898,0.22::899>, <0.20::874,0.72::875,0.08::876>), (<0.07::531,0.82::532,0.11::533>, <0.37::896,0.54::897,0.09::898>, <-0.00::874,0.77::875,0.23::876>), (<0.12::24,0.72::25,0.16::26>, <0.01::895,0.78::896,0.21::897>, <0.34::874,0.66::875,-0.00::876>), (<0.19::30,0.74::31,0.07::32>, <0.20::894,0.61::895,0.19::896>, <0.00::874,0.99::875,0.01::876>), (<0.20::42,0.77::43,0.03::44>, <0.02::893,0.95::894,0.03::895>, <0.08::874,0.73::875,0.19::876>), (<0.05::573,0.85::574,0.10::575>, <0.08::892,0.91::893,0.01::894>, <0.06::874,0.81::875,0.13::876>), (<0.14::760,0.53::761,0.33::762>, <0.15::891,0.65::892,0.20::893>, <0.13::874,0.66::875,0.21::876>), (<0.05::575,0.65::576,0.30::577>, <0.01::890,0.82::891,0.17::892>, <0.12::874,0.66::875,0.22::876>), (<0.00::290,0.94::291,0.06::292>, <0.02::889,0.95::890,0.03::891>, <0.12::874,0.86::875,0.02::876>), (<0.14::916,0.84::917,0.02::918>, <0.02::888,0.95::889,0.03::890>, <0.01::874,0.54::875,0.45::876>); \\
        row 3: (<0.21::531,0.77::532,0.02::533>, <0.01::496,0.88::497,0.11::498>, <0.07::830,0.62::831,0.31::832>), (<0.20::24,0.79::25,0.01::26>, <0.19::495,0.62::496,0.19::497>, <0.06::830,0.92::831,0.02::832>), (<0.17::30,0.56::31,0.27::32>, <0.27::494,0.64::495,0.09::496>, <0.02::830,0.98::831,0.00::832>), (<0.00::42,0.98::43,0.02::44>, <0.38::493,0.58::494,0.04::495>, <0.19::830,0.53::831,0.28::832>), (<0.07::573,0.52::574,0.41::575>, <0.01::492,0.99::493,0.00::494>, <0.01::830,0.81::831,0.18::832>), (<0.26::760,0.55::761,0.19::762>, <0.13::491,0.83::492,0.04::493>, <0.05::830,0.82::831,0.13::832>), (<0.47::575,0.52::576,0.01::577>, <0.15::490,0.59::491,0.26::492>, <0.16::830,0.81::831,0.03::832>), (<0.03::290,0.82::291,0.15::292>, <0.29::489,0.52::490,0.19::491>, <0.03::830,0.85::831,0.12::832>), (<0.08::916,0.81::917,0.11::918>, <0.05::488,0.83::489,0.12::490>, <0.09::830,0.64::831,0.27::832>), \\
        \vspace{-1mm}
        Answer set:\\
        Answer \#0: (<0.06::289,0.83::290,0.11::291>, <0.00::487,1.00::488,0.00::489>, <0.03::874,0.82::875,0.15::876>)\\
        Answer \#1: (<0.01::850,0.78::851,0.21::852>, <0.00::487,0.99::488,0.01::489>, <0.08::874,0.85::875,0.07::876>)\\
        Answer \#2: (<0.03::289,0.57::290,0.40::291>, <0.15::450,0.75::451,0.10::452>, <0.15::830,0.62::831,0.23::832>)\\
        Answer \#3: (<0.06::289,0.52::290,0.42::291>, <0.03::487,0.92::488,0.05::489>, <0.31::830,0.61::831,0.08::832>)\\
        Answer \#4: (<0.02::850,0.95::851,0.03::852>, <0.16::450,0.63::451,0.21::452>, <0.20::874,0.52::875,0.28::876>)\\
        Answer \#5: (<0.02::850,0.86::851,0.12::852>, <0.18::487,0.80::488,0.02::489>, <0.14::830,0.79::831,0.07::832>)\\
        Answer \#6: (<0.01::289,0.96::290,0.03::291>, <0.38::450,0.59::451,0.03::452>, <0.08::874,0.68::875,0.24::876>)\\
        Answer \#7: (<0.08::850,0.62::851,0.30::852>, <0.15::450,0.82::451,0.03::452>, <0.09::830,0.87::831,0.04::832>)\\
    \bottomrule
    \end{tabular}
    \caption{Example prompt for the I-RAVEN-X task with smooth distributions.}
    \label{fig:iraven_prompt_unc}
\end{table}

\clearpage
\section{Comparison between OpenAI o3-mini and o1}
\label{appx:o1}
This Appendix presents a small ablation study on two different closed-source \glspl*{lrm}, OpenAI o1 and OpenAI o3-mini. 
The goal of these experiments was to measure the difference, if any, in the reasoning capabilities of the o3-mini model compared to its bigger, more expensive predecessor.
We restricted the size of the test set to 100 test examples for both I-RAVEN and I-RAVEN-X.
The results, presented in Table \ref{tab:o1ablation}, show that the two models achieve roughly comparable performance on both I-RAVEN and I-RAVEN-X, with o3-mini being consistently slightly less accurate than o1.
However, o1 is also considerably more expensive compared to o3: o1 is priced at \$15 and \$60 per million input and output tokens, respectively, while o3-mini costs only \$1.1 and \$4.4 per million input and output tokens (approximately $14\times$ less expensive).
Hence, we opt to use only o3-mini in the full evaluation.

\begin{table}[h!]
\centering
\begin{tabular}{lccccccc}
\toprule
\multirow{4}{*}{\textbf{Model}} & \multirow{4}{*}{\textbf{Setting}} & \multicolumn{2}{c}{\textbf{I-RAVEN}} & \multicolumn{4}{c}{\textbf{I-RAVEN-X}} \\
\cmidrule(r){3-4}\cmidrule(r){5-8}
& & \multicolumn{2}{c}{\textbf{Range 10}}  & \multicolumn{2}{c}{\textbf{Range 100}}      & \multicolumn{2}{c}{\textbf{Range 1000}}    \\
& & Task & Arithm. &  Task & Arithm. & Task & Arithm.  \\
\midrule
OpenAI o1       & Entangled &   88.0 & 79.7 & 86.0 & 68.2 & 86.0 & 68.2 \\
OpenAI o3-mini  & Entangled &   86.6 &  81.4 &  84.0 & 63.6 & 81.0 & 60.8 \\
\bottomrule
\end{tabular}
\caption{Task and arithmetic accuracy (\%) comparison of two different \glspl*{lrm} on a subset of 100 test examples of I-RAVEN and I-RAVEN-X.}
\label{tab:o1ablation}
\end{table}

\end{document}